# THE ORTHOGLIDE:
# KINEMATICS AND WORKSPACE ANALYSIS

A. Pashkevich

*Robotic Laboratory, Belarusian State University of Informatics and Radioelectronics*
*6 P. Brovka St., Minsk, 220027, Belarus*
E-mail: pap@ bsuir.unibel.by

D. Chablat, P. Wenger

*Institut de Recherche en Communications et Cybernétique de Nantes*
*1 rue de la Noë, BP 92 101, 44321 Nantes, France*
E-mails:Damien.Chablat@irccyn.ec-nantes.fr, Philippe.Wenger@irccyn.ec-nantes.fr

**Abstract**  The paper addresses kinematic and geometrical aspects of the Orthoglide, a three-DOF parallel mechanism. This machine consists of three fixed linear joints, which are mounted orthogonally, three identical legs and a mobile platform, which moves in the Cartesian *x-y-z* space with fixed orientation. New solutions to solve inverse/direct kinematics are proposed and a detailed workspace analysis is performed taking into account specific joint limit constraints.

## 1. Introduction

For two decades, parallel manipulators attract the attention of more and more researchers who consider them as valuable alternative design for robotic mechanisms (Asada et al, 1986, Fu et al., 1987, Craig, 1989). As stated by a number of authors (Tsai, 1999), conventional serial kinematic machines have already reached their dynamic performance limits, which are bounded by high stiffness of the machine components required to support sequential joints, links and actuators. Thus, while having good operating characteristics (large workspace and high flexibility), serial manipulators have disadvantages of low precision, low stiffness and low power. Also, they are generally operated at low speed to avoid excessive vibration and deflection.

Conversely, parallel kinematic machines offer essential advantages over their serial counterparts (lower moving masses and higher rigidity) that obviously should lead to higher dynamic capabilities. However, most existing parallel manipulators have limited and complicated workspace with singularities, and highly non-isotropic input/output relations (Angeles, 2002). Hence, the performances may significantly vary over the workspace and depend on the direction of the motion, which is a serious disadvantage for machining applications. Research in the field of parallel manipulators began with the Stewart-platform used in flight simulators (Stewart, 1965). Many such structures have been investigated since then, which are composed of six linearly actuated legs with different combinations of link-to-platform connections (Merlet, 2000). In recent years, several new kinematic structures have been proposed that possess higher isotropy. In particular, a 3-dof translational mechanism with gliding foot points was found in three separate works to be fully isotropic throughout the Cartesian workspace (Carricato et all, 2002 and Kong et all, 2002). It consists of a mobile platform, which is connected to three orthogonal linear drives through three identical planar 3-revolute jointed serial chains. Although this manipulator behaves like a conventional Cartesian machine, bulky legs are required to assure stiffness because these legs are subject to bending.

In this paper, the Orthoglide manipulator proposed by Wenger et all, 2000, is studied. As follows from previous research, this manipulator has good kinetostatic performances and some technological advantages, such as (i) symmetrical design consisting of similar 1-d.o.f. joints; (ii) regular workspace shape properties with bounded velocity amplification factor; and (iii) low inertia effects (Chablat et all, 2003). This article analyses the kinematics and the workspace of the Orthoglide. Section 2 describes the Orthoglide geometry. Section 3 proposes new solutions for its inverse and direct kinematics. Sections 4, 5 present a detailed analysis of the workspace and jointspace respectively. Finally, Section 6 summarises the main contributions of the paper.

## 2. Manipulator geometry

The kinematic architecture of the Orthoglide is shown in Fig. 1. It consists of three identical kinematic chains that are formally described as $PRP_aR$, where $P$, $R$ and $P_a$ denote the prismatic, revolute, and parallelogram joints respectively. The mechanism input is made up by three actuated orthogonal prismatic joints. The output body is connected to the prismatic joints through a set of three kinematic chains. Each chain includes a parallelogram, so that the output body is restricted to translational movements. To get the Orthoglide kinematic equations, let us locate the reference frame at the intersection of the prismatic joint axes and align the coordinate axis with them (Fig. 2), following the "right-hand" rule. Let us also denote the input vector of the prismatic joints variables as $\boldsymbol{\rho} = (\rho_x, \rho_y, \rho_z)$ and the output position vector of the tool centre point as $\mathbf{p} = (p_x, p_y, p_z)$. Taking into account obvious properties of the parallelograms, the Orthoglide geometrical model can be presented in a simplified form, which consists of three bar links connected by spherical joints to the tool centre point at one side and to the corresponding prismatic joints at another side.

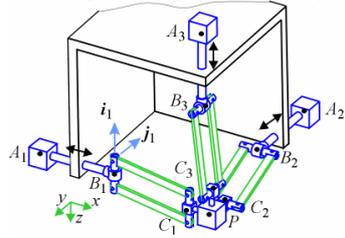 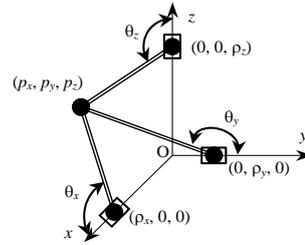

Fig. 1. Orthoglide architecture    Fig. 2. Orthoglide geometrical model

Using this notation, the kinematic equations of the Orthoglide can be written as follows

$$(p_x - \rho_x)^2 + p_y^2 + p_z^2 = L^2, \; p_x^2 + (p_y - \rho_y)^2 + p_z^2 = L^2, \; p_x^2 + p_y^2 + (p_z - \rho_z)^2 = L^2 \quad (1)$$

where $L$ is the length of the parallelogram principal links and the "zero" position $\mathbf{p}_0 = (0, 0, 0)$ corresponds to the joints variables $\boldsymbol{\rho}_0 = (L, L, L)$. It should be stressed that the Orthoglide geometry and relevant manufacturing technology impose the following constraints on the joint variables

$$0 < \rho_x \leq 2L \; ; \; 0 < \rho_y \leq 2L \; ; \; 0 < \rho_z \leq 2L, \quad (2)$$

which essentially influence on the workspace shape. While the upper bound ($\rho \leq 2L$) is implicit and obvious, the lower one ($\rho > 0$) is caused by practical reasons, since safe mechanical design encourages avoiding risk of simultaneous location of prismatic joints in the same point of the Cartesian workspace (here and in the following sections, while referring to symmetrical constraints are subscript omitted, *i.e.* $\rho \in \{\rho_x, \rho_y, \rho_z\}$).

### 3. Orthoglide Kinematics

### 3.1 Inverse kinematics

For the inverse kinematics, the position of the end-point ($p_x, p_y, p_z$) is treated as known and the goal is to find the joint variables ($\rho_x, \rho_y, \rho_z$) that yield the given location of the tool. Since in the general case the inverse kinematics can produce several solutions corresponding to the same tool location, the solutions must be distinguished with respect to the algorithm "branch". For instance, if the aim is to generate a sequence of points to move the tool along an arc, care must be taken to avoid branch switching during motion, which may cause inefficient (or even impossible) manipulator motions. Moreover, leg singularities may occur at which the manipulator loses degrees of freedom and the joint variables become linearly dependent. Hence, the complete investigation of the Orthoglide kinematics must cover all the above-mentioned topics.

From the Orthoglide geometrical model (1), the inverse kinematic equations can be derived in a straightforward way as:

$$\rho_x = p_x + s_x \sqrt{L^2 - p_y^2 - p_z^2} \quad \rho_y = p_y + s_y \sqrt{L^2 - p_x^2 - p_z^2} \quad \rho_z = p_z + s_z \sqrt{L^2 - p_x^2 - p_y^2} \quad (3)$$

where $s_x$, $s_y$, $s_z$ are the branch (or configuration) indices that are equal to ±1. It is obvious that (3) yields eight different branches of the inverse kinematic algorithm, which will be further referred to as *PPP*, *MPP*...*MMM* following the sign of the corresponding index (i.e. the notation *MPP* corresponds to the indices $s_x = -1; s_y = +1; s_z = +1$). The geometrical meaning of these indices is illustrated by Fig. 2, where $\theta_x$, $\theta_y$, $\theta_z$ are the angles between the bar links and the corresponding prismatic joint axes. It can be proved that $s = 1$ if $\theta \in (90°, 180°)$ and $s = -1$ if $\theta \in (0°, 90°)$. The branch transition ($\theta = 90°$) corresponds to the serial singularity (where the leg is orthogonal to the relevant translational axis and the input joint motion does not produce the end-point displacement). It is obvious that if the inverse kinematic solution exists, then the target point ($p_x$, $p_y$, $p_z$) belongs to a volume bounded by the intersection of three cylinders

$$C_L = \left\{ \mathbf{p} \mid p_x^2 + p_y^2 \le L^2; \ p_x^2 + p_z^2 \le L^2; \ p_y^2 + p_z^2 \le L^2 \right\} \quad (4)$$

that guarantees non-negative values under the square roots in (3). However, it is not sufficient, since the lower joint limits (2) impose the following additional constraints

$$p_x > -s_x \sqrt{L^2 - p_y^2 - p_z^2} \ ; \ p_y > -s_y \sqrt{L^2 - p_x^2 - p_z^2} \ ; \ p_z > -s_z \sqrt{L^2 - p_x^2 - p_y^2} \quad (5)$$

which reduce a potential solution set. For example, it can be easily computed that for the "zero" workspace point $\mathbf{p}_0 = (0, 0, 0)$, the inverse kinematic equations (3) give eight solutions $\boldsymbol{\rho} = (\pm L, \pm L, \pm L)$ but only one of them is feasible. To analyse in details the influence of the joint constraints impact, let us start from separate a study of the inequalities (5) and then summarise results for all possible combinations of the three configuration indices. If $s_x = 1$, then consideration of two cases, $p_x > 0$ and $p_x \le 0$, yields the following workspace set satisfying the constraint $\rho_x > 0$

$$W_L^{+x} = \left\{ \mathbf{p} \in C_L \mid p_x > 0 \right\} \cup \left\{ \mathbf{p} \in C_L \mid p_x \le 0; p_x^2 + p_y^2 + p_z^2 < L^2 \right\} \quad (6)$$

which consists of two fractions (½ of the cylinder intersection denoted $C_L$ and ½ of the sphere whose geometric center is (0,0,0) and radius is *L*).

If $s_x = -1$, then the second case $p_x < 0$ does not give any solution and the joint constraint $\rho_x > 0$ is expressed in the workspace as

$$W_L^{-x} = \{\mathbf{p} \in C_L \mid p_x \geq 0;\ p_x^2 + p_y^2 + p_z^2 > L^2\}. \qquad (7)$$

The latter defines a solid bounded by three cylindrical surfaces and the sphere. The remaining constrains $\rho_y > 0$ and $\rho_z > 0$ can be derived similarly, which differ from (6), (7) by subscripts only.

Then, there can be found intersection of the obtained sets for different combinations of the configuration indices. It can be easily proved that the case "PPP" yields

$$W_L^{PPP} = \{\mathbf{p} \in C_L \mid p_x, p_y, p_z > 0\} \cup \{\mathbf{p} \in C_L \mid p_x^2 + p_y^2 + p_z^2 < L^2\} \qquad (8)$$

while the remaining cases give

$$W_L^{MPP} = \ldots = W_L^{MMM} = \{\mathbf{p} \in C_L \mid p_x, p_y, p_z > 0\} \cap \{\mathbf{p} \in C_L \mid p_x^2 + p_y^2 + p_z^2 > L^2\} \qquad (9)$$

Expressions (8) and (9) can be put in the form

$$W_L^{PPP} = S_L \cup G_L; \qquad W_L^{MPP} = \ldots = W_L^{MMM} = G_L \qquad (10)$$

where $S_L = \{\mathbf{p} \in C_L \mid p_x^2 + p_y^2 + p_z^2 < L^2\}$;
$G_L = \{\mathbf{p} \in C_L \mid p_x, p_y, p_z > 0;\ p_x^2 + p_y^2 + p_z^2 > L^2\}$; $S_L \cap G_L = \varnothing$.

Therefore, for the considered positive joint limits (2), the existence of the inverse kinematic solutions may be summarised as follows (i) inside the sphere $S_L$ there exist exactly one inverse kinematic solution *PPP* with positive configuration indices $s_x$, $s_y$, $s_z$, (ii) outside the sphere $S_L$, but within the positive part of the cylinder intersection $C_L$, there exist 8 solutions of the inverse kinematics (*PPP, MPP, … MMM*) corresponding to all possible combinations of the configuration indices $s_x$, $s_y$, $s_z$. These conclusions may be illustrated when $L = 1$ by numerical examples. If the target point $\mathbf{p}=(-0.5, 0.4, 0.3)$ is within the sphere $S_L$, then the joint coordinates must be taken from the sets $\rho_x \in \{0.37, -1.37\}$, $\rho_x \in \{1.21, -0.41\}$, $\rho_x \in \{1.07, -0.47\}$, which allow only one positive combination. In contrast, for the target point $\mathbf{p}=(0.7, 0.7, 0.7)$, which is outside the sphere, the inverse kinematics yields solutions with two positive values $\rho_x, \rho_y, \rho_z \in \{0.84, 0.56\}$ that allow 8 positive combinations of the joint variables. An interesting feature is that intermediate cases (with 2 or 4 feasible solutions) are not possible.

### 3.2 Direct kinematics

For the direct kinematics, the values of the joint variables ($\rho_x$, $\rho_y$, $\rho_z$) are known and the goal is to find the tool centre point location ($p_x$, $p_y$, $p_z$) that corresponds to the given joint positions. While, in general, the inverse kinematics of parallel mechanisms is straightforward, the direct kinematics is usually very complex. The Orthoglide has the advantage leave an analytical direct kinematics. Like for the previous section, the solutions must be distinguished with respect to the algorithm "branch" that should be also defined both geometrically and algebraically, via a configuration index.

To solve the system (1) for $p_x$, $p_y$, $p_z$, first, let us derive linear relations between the unknowns. By subtracting three possible pairs of the equations (1), we leave

$$2\rho_x p_x - 2\rho_y p_y = \rho_x^2 - \rho_y^2 \ , \ 2\rho_x p_x - 2\rho_z p_z = \rho_x^2 - \rho_z^2 \ , \ 2\rho_y p_y - 2\rho_z p_z = \rho_y^2 - \rho_z^2 \qquad (11)$$

As follows from these expressions, the relation between $p_x$, $p_y$, $p_z$ may be presented as

$$p_x = \rho_x/2 + t/\rho_x; \ p_y = \rho_y/2 + t/\rho_y; \ p_z = \rho_z/2 + t/\rho_z, \qquad (12)$$

where $t$ is an auxiliary scalar parameter. From a geometrical point of view, the expression (12) defines the set of equidistant points for the prismatic joint centres (Fig. 6). Also, it can be easily proved that the full set of equidistant points is the line perpendicular to $\Pi$ and passing through $(\rho_x, \rho_y, \rho_z)/2$, where.

$$\Pi = \{\ \mathbf{p} \mid \ p_x/\rho_x + p_y/\rho_y + p_z/\rho_z = 1\ \} \qquad (13)$$

After substituting (12) into any of the equations (1), the direct kinematic problem is reduced to the solution of a quadratic equation in the auxiliary variable $t$,

$$At^2 + Bt + C = 0, \qquad (14)$$

where $A = (\rho_x \rho_y)^2 + (\rho_x \rho_z)^2 + (\rho_y \rho_z)^2$, $B = (\rho_x \rho_y \rho_z)^2$, $C = (\rho_x^2 + \rho_y^2 + \rho_z^2 - 4L^2)B/4$.
The quadratic formula yields two solutions

$$t = (-B + m\sqrt{B^2 - 4AC})/(2A); \ m = \pm 1 \qquad (15)$$

that geometrically correspond to different locations of the target point $P$ (see Fig. 6) with respect to the plane passing through the prismatic joint centres (it should be noted that the intersection point of the plane and the set of equidistant point corresponds to $t_0 = -B/(2A)$). Hence, the Orthoglide direct kinematics is solved analytically, via the quadratic formula (14) for the auxiliary variable $t$ and its substitution into expressions (12). The direct kinematic solution exists if and only if the joint variables satisfy the inequality $B^2 \geq 4AC$, which defines a closed region in the joint variable space

$$\Re_L = \{\mathbf{\rho} \mid \left(\rho_x^2 + \rho_y^2 + \rho_z^2 - 4L^2\right)\left(\rho_x^{-2} + \rho_y^{-2} + \rho_z^{-2}\right) \leq 1\} \qquad (16)$$

Taking into account the joint limits (2), the feasible joint space may be presented as $\Re_L^+ = \{\mathbf{\rho} \in \Re_L \mid \rho_x, \rho_y, \rho_z > 0\}$. Therefore, for the considered positive joint limits (2), the existence of the direct kinematic solutions may be summarised as follows: (i) inside the region $\Re_L^+$, there exist exactly two direct kinematic solutions, which differ by the target point location relative to the plane $\Pi$ (Fig. 7a). (ii) On the border of the region $\Re_L^+$ located inside the first octant, there exist a single direct kinematic solution, which corresponds to the "flat" manipulator configuration, where both the target point and prismatic joint centres belong to the plane $\Pi$ (Fig. 7b).

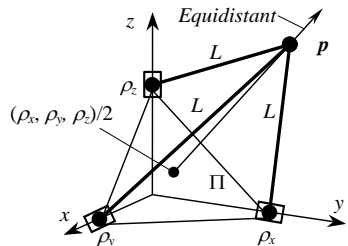
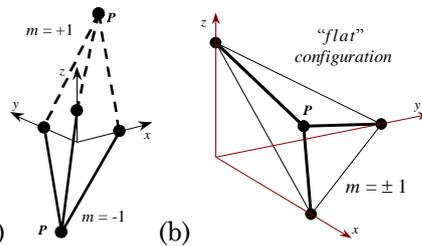

Fig. 6. Geometrical solution of the direct kinematics

Fig. 7. Double (a) and single (b) solutions of the direct kinematics

These conclusions may be illustrated by the following numerical examples (for the "unit" manipulator, $L=1$). Since the joint variables $\rho_x=\rho_y=\rho_z=0.3$ are within $\Re_L^+$, then the end-point coordinates are either $p_x=p_y=p_z=-0.46$ or $p_x=p_y=p_z=-0.66$. In contrast, for the joint variables $\rho_x=\rho_y=\rho_z=\sqrt{1.5}$, which are exactly on the surface $\Re_L^+$, the direct kinematics yields a single solution $p_x=p_y=p_z=\sqrt{1/6}$ corresponding to the "flat" configuration (see Fig. 7b).

### 3.3 Configuration indices

As follows from the previous sub-sections, both the inverse and direct kinematics of the Orthoglide may produce several solutions. The problem is how to define numerically the *configuration indices*, which allow choosing among the corresponding algorithm branches.

For the inverse kinematics, when the configuration is defined by the angle between the leg and the corresponding prismatic joint axis, the decision equations for the configuration indices are trivial:

$$s_x = \operatorname{sgn}(\rho_x - p_x);\ \ s_y = \operatorname{sgn}(\rho_y - p_y);\ \ s_z = \operatorname{sgn}(\rho_z - p_z)$$

Geometrically, $s > 0$ means that (see Fig. 2), $\theta_x, \theta_y, \theta_z \in\, ]\pi/2\ \ 3\pi/2[$. For the direct kinematics, the configuration is defined by the end-point location relative to the plane that passes through the prismatic joint centres (see Figs. 6-7). Hence, the decision equation may be derived by analysing the dot-product of the plane normal vector $\left(\rho_x^{-1},\ \rho_y^{-1},\ \rho_z^{-1}\right)$ and the vector directed along any of the bar links (for instance, $(p_x-\rho_x,\ p_y,\ p_z)$ for the first link: $m=\mathrm{sgn}\left(p_y/\rho_x + p_y/\rho_y + p_z/\rho_z -1\right)$ which is equivalent to $m=\mathrm{sgn}\left(p_x\rho_y\rho_z + \rho_x p_y\rho_z + \rho_x\rho_y p_z - \rho_x\rho_y\rho_z\right)$ for the positive joint limits.

It should be stressed that the feasible solutions for the inverse/direct kinematics, located in the neighbourhood of the "zero" point, have the configuration indices $s_x=s_y=s_z=+1$ and $m=-1$.

## 4. Workspace analysis

As follows from Eq. (10), Orthoglide workspace $W_L$ is composed of two fractions: (i) the sphere $S_L$ of radius $L$ and centre point (0, 0, 0), and (ii) the thin non-convex solid $G_L$, which is located in the first octant and bounded by the surfaces of the sphere $S_L$ and the cylinder intersection $C_L$. It can be proved that the volume of $C_L$, $S_L$ and $W_L$ is defined by the expressions

$$Vol(C_L)=8\left(2-\sqrt{2}\right)L^3,\ Vol(G_L)=\left(2-\sqrt{2}-\pi/6\right)L^3,\ Vol(W_L)=\left(2+7\pi/6-\sqrt{2}\right)L^3 \quad (17)$$

As follows from (17), the Orthoglide with the joint limits (2) uses about 53% of the workspace $V_{PPP}$ of its serial counterpart (a Cartesian PPP machine with $2L\times 2L\times 2L$ workspace). Also, the volume of $G_L$ ($0.062\,L^3$) is insignificant in comparison to the volume of sphere $S_L$ ($4.189\,L^3$), which is equal to 52% of $V_{PPP}$. On the other hand, releasing the lower joint limit ($\rho>0$) leads to an increases the workspace volume of up to 59% of $V_{PPP}$ only, since the volume of the workspace is, then, equal to $C_L$. The mutual location of $G_L$ and $S_L$ (and their size ratio) may be also evaluated by the intersection points of the first octant bisector. In particular, for the sphere $S_L$ the bisector intersection point is located at distance $1/\sqrt{3}\approx 0.58$ from the origin, while for the solid $G_L$ the corresponding distance is $1/\sqrt{2}\approx 0.71$ (assuming that $L=1$). Moreover, $G_L$ touches the sphere by its circular edges, which are located on the borders of the first octant.

## 5. Joint Space Analysis

The properties of the *feasible* jointspace are essential for the Orthoglide control, in order to avoid impossible combinations of the prismatic joint variables $\rho_x$, $\rho_y$, $\rho_z$, which are generated by the control system and are followed by the actuators. For serial manipulators, this problem does not usually exist because the jointspace is bounded by a parallelepiped and mechanical limitations of the joint values may be verified easily and independently. For parallel manipulators, however, we needs to check both (i) separate input coordinates (to satisfy the joint limits), and (ii) their combinations that must be feasible to produce a direct kinematic solution.

As follows from Sub-Section 3.3, the Orthoglide jointspace $\mathfrak{R}_L^+$ is located within the first octant and is bounded by a surface, which corresponds to a single solution of the direct kinematics. Therefore, the jointspace boundary is defined by the relation $B^2 = 4AC$ (see equation (14)), which may be rewritten as

$$\left(\rho_x^2 + \rho_y^2 + \rho_z^2 - 4L^2\right)\left(\rho_x^{-2} + \rho_y^{-2} + \rho_z^{-2}\right) = 1 \qquad (18)$$

and solved for $\rho_x$ assuming that $\rho_y, \rho_z$ are known, $D\rho_x^4 + DE\rho_x^2 + E = 0$, where $D = \rho_y^{-2} + \rho_z^{-2}$; $E = \rho_y^2 + \rho_z^2 - 4L^2$. However, this equation is non-symmetrical with respect to $\rho_x, \rho_y, \rho_z$ and, therefore, is not convenient the real-time control. An alternative way to obtain the jointspace boundary, which is more computationally efficient, is based on the conversion from Cartesian to spherical coordinates, $\rho_x = e_x t$, $\rho_y = e_y t$, $\rho_z = e_z t$, where $t \geq 0$ is the length of the vector $\boldsymbol{\rho}$, and ($e_x, e_y, e_z$) are the components of the unit direction vector, which are expressed via two angles $\varphi, \theta$ with $e_x = \cos\varphi \cos\theta$, $e_y = \cos\varphi \sin\theta$, $e_z = \sin\varphi$, where $\varphi, \theta \in \left]0, \pi/2\right]$. For such a notation, the original equation (18) is transformed into a linear equation for $t^2$, $(F-1)t^2 = 4L^2 F$, $F = e_x^{-2} + e_y^{-2} + e_z^{-2}$, with an obvious solution $t = 2L\sqrt{F/(F-1)}$. As follows from its analyses, the bounding surface is close to the 1/8th of the sphere $S_{2L}$. At the edges, which are exactly quarters of the circles of the radius $2L$, the surface touches the sphere. However, in the middle, the surface is located out of the sphere. In particular, the intersection point of the first octant bisector is located at the distance $\sqrt{3/2} \approx 1.22$ from the coordinate system origin for the jointspace border and at the distance $2/\sqrt{3} \approx 1.15$ for the sphere $S_{2L}$ (assuming $L=1$).

## 6. Conclusions

This article focuses on the kinematics and workspace analysis of the Orthoglide, a 3-DOF parallel mechanism with a kinematic behaviour close to the conventional Cartesian machine taking into account the specific manufacturing constraints in the joint variables. We proposed a formal definition of the configuration indices and developed new simple analytical expressions for the Orthoglide inverse/direct kinematics. It was proved that, for the considered joint limits, the Orthoglide workspace is composed of two fractions, the sphere and a thin non-convex solid in which there are 1 and 8 inverse kinematic solutions, respectively. The total workspace volume comprises about 53% of the corresponding serial machine workspace, where over 52% belongs to the sphere (for comparison, releasing of the joint limits yields to an increase of up to 59% in the workspace volume). It was also shown, that the Orthoglide jointspace is bounded by surface with circular edges, which is more convex than the sphere but is rather close to it. These results can be further used for the optimisation of the Orthoglide parameters, which is the subject of our future research work.